\documentclass[10pt,twocolumn,letterpaper]{article}

\usepackage[pagenumbers]{cvpr}      % To produce the REVIEW version
\definecolor{cvprblue}{rgb}{0.21,0.49,0.74}
\usepackage[pagebackref,breaklinks,colorlinks,allcolors=cvprblue]{hyperref}

\def\confName{CVPR}
\def\confYear{2027}

\usepackage{algorithm}
\usepackage{algpseudocode}
\usepackage{amsmath,amssymb}

\usepackage{booktabs}
\usepackage{multirow}
\usepackage{array}
\usepackage{amsmath,amssymb}
\usepackage{graphicx}

\title{Dually Constrained Parameter Projection for Continual Learning}

\author{%
  \textbf{Bowyn Tan}$^{1,2}$ \quad
  \textbf{Zhilong Zheng}$^{1}$ \quad
  \textbf{Yang Guan}$^{1}$\thanks{Corresponding authors.} \quad
  \textbf{Liming Xiao}$^{2}$ \quad
  \textbf{Liye Tang}$^{2}$ \\[0.5ex]
  \textbf{Xiao Li}$^{2}$ \quad
  \textbf{Naizheng Wang}$^{2}$ \quad
  \textbf{Keqiang Li}$^{1}$ \quad
  \textbf{Shengbo Eben Li}$^{1}$\footnotemark[1] \\[1ex]
  $^1$Tsinghua University \quad
  $^2$Meituan
}

\begin{document}
\maketitle
\begin{abstract}
Continual learning requires a model to learn multiple new tasks sequentially while retaining performance on previous tasks. Projecting model parameters onto the Jacobian null space of previous tasks after unconstrained training on a new task has emerged as an effective paradigm. However, unidirectional projection merely minimizes the parameter distance to the new-task optimum, overlooking the new-task loss landscape. To address this limitation, we propose DUET, a bidirectional projection algorithm that imposes Jacobian null-space constraints from both previous and new tasks during post-training projection. Specifically, DUET formulates the post-training adjustment as a dually constrained optimization problem. It minimizes the distance between parameters before and after training on a new task, subject to loss preserving constraints on both tasks.  To solve the optimization problem, we develop an alternating bidirectional projection algorithm, whose convergence is guaranteed by the contraction mapping theorem.  For reliable multi-step projection, we rectify existing step size selection criteria via an effective-rank aware formulation, overcoming their failure under rank-deficient Jacobians. Extensive experiments demonstrate that DUET consistently outperforms existing unidirectional projection methods and achieves state-of-the-art performance on multiple datasets.
\end{abstract}

\begin{figure*}[t]
    \centering
    \includegraphics[width=0.98\textwidth]{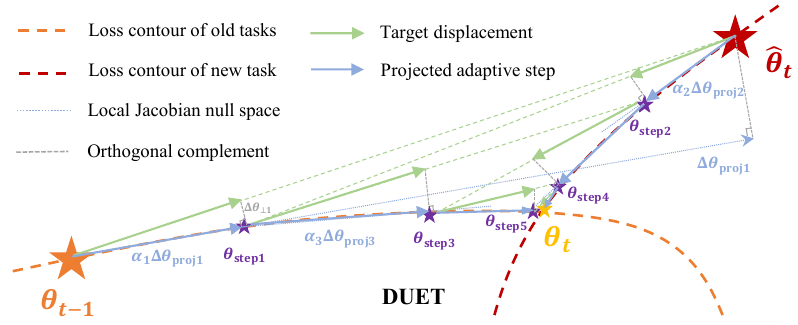}
    \caption{Geometric illustration of alternating bidirectional projection in DUET.}
    \label{fig:DUET_mechanism}
\end{figure*}    
\section{Introduction}
\label{sec:intro}

As deep neural networks continue to expand their capabilities across diverse tasks, the ability to continuously acquire new knowledge without retraining from scratch has become increasingly crutial for applications across computer vision~\cite{CVCL}, large language models~\cite{LLMCL}, autonomous driving~\cite{guan2026enhanced}, sequential decision-making~\cite{guan2027beyond}, and reinforcement learning~\cite{guan2021direct}. \textbf{Continual learning (CL)} requires a model to learn a sequence of tasks while retaining knowledge from previous tasks~\cite{PARISI201954}. Learning new tasks often causes catastrophic forgetting, where updates to model parameters interfere with previously acquired knowledge~\cite{kemker2018measuring}. Existing methods attempt to strike a balance between previous-task stability and new-task plasticity across different training stages~\cite{hou2026continual}. Parameter efficient fine-tuning (PEFT) initialization methods~\cite{meng2024pissa, yang2024corda, tang2026put, wang2024milora} initializes and freezes~\cite{liang2024inflora, luo2026keeplora, tang2026put} one of the low-rank factors of LoRA within the orthogonal complement of historical activations, which ensures unchanged layer output of previous tasks but is overly restrictive and limits new-task plasticity. During-training methods mitigate forgetting through regularization penalties~\cite{kirkpatrick2017overcoming, krukowski2025intact, liu2026octopus} or gradient projection methods. The latter suppress interference by projecting gradients at each optimization step, either onto the orthogonal complement of historical activations~\cite{saha2021gradient, liang2023adaptive} or onto the Jacobian null space of previous tasks~\cite{farajtabar2020orthogonal, zhan2025bicriteria, Paz2017GEM, Arslan2019AGEM, Tekmen2025IGEM}. But they interfere with training dynamics and limits new-task plasticity. Post-hoc rectification methods perform unconstrained training on a new task and subsequently adjust model parameters via parameter pruning~\cite{huang2025mitigating}, task vectors~\cite{ilharco2022editing, yadav2023ties}, low-rank compression~\cite{vasani2025phlora, SVDLLM, SVDQuant, DobiSVD}, or post-hoc parameter projection~\cite{zheng2026past, hsu2024safelora, fang2025alphaedit}. By granting uninhibited optimization freedom during training, post-hoc rectification methods have a higher plasticity ceiling on new tasks with minimal computation, emerging as a highly promising paradigm. 

Among post-hoc rectification methods, post-hoc parameter projection approaches~\cite{zheng2026past, hsu2024safelora, fang2025alphaedit} provide more rigorous theoretical guarantees against forgetting. Projecting updates onto the activation null space ensures unchanged layer-level outputs~\cite{hsu2024safelora, fang2025alphaedit}, but imposes overly restrictive constraints that severely limit new-task plasticity. In contrast, projecting parameter updates onto the Jacobian null space of previous tasks~\cite{zheng2026past} relaxes this requirement to a first-order loss invariance constraint, which effectively balances stability and plasticity, demonstrating superior performance and promising potential.

However, existing post-hoc projection methods~\cite{zheng2026past, hsu2024safelora, fang2025alphaedit} merely utilizing unidirectional projection to enforces loss preservation on previous tasks. Unidirectional projection minimizes the Euclidean distance between the adjusted parameters and the new-task optimum, treating parameter distance to new-task optimum as a proxy for lower new-task loss. It neglects the new-task loss landscape, inevitably steering the model into suboptimal parameter regions with degraded new-task performance.

Meanwhile, projecting parameter updates onto the Jacobian null space~\cite{zheng2026past} aims to ensure first-order loss invariance on old tasks. Since this invariance holds only locally, multi-step projection requires an adaptive step size, typically guided by a cosine similarity criterion to verify local linearity~\cite{zheng2026past}. This criterion measures the alignment between the Jacobian subspaces before and after a candidate update to detect geometric distortion. However, we find that when the Jacobians suffer from effective-rank deficiency, the cosine similarity is systematically underestimated, remaining low even at a zero step size. This underestimation repeatedly triggers false step rejections, collapses the step size to zero, and prematurely stalls the multi-step projection process.

To address these limitations, we advance the post-hoc unidirectional Jacobian null-space projection method in \textbf{JANUS}~\cite{zheng2026past} and propose \textbf{DU}ally constrained param\textbf{E}ter projec\textbf{T}ion (\textbf{DUET}), a bidirectional null-space projection algorithm that incorporates Jacobian null-space constraints from both previous and new tasks. Specifically, DUET formulates the post-training parameter adjustment as a constrained optimization problem that minimizes the distance between the parameters before and after training on a new task, while satisfying the loss-preserving constraints of both tasks. To solve the constrained optimization problem, we develop an alternative bidirectional Jacobian null-space projection algorithm (See Fig.~\ref{fig:DUET_mechanism}), whose convergence is guaranteed by the contraction mapping theorem. Furthermore, we propose an effective-rank aware cosine criterion that measures Jacobian subspace alignment only within informative directions. It identifies the dominant directions via spectral truncation~\cite{doi:10.1137/1.9781421407944, doi:10.1137/1.9780898719697}, computing the cosine similarity exclusively over the dominant singular subspace. This enables more reliable adaptive step-size adjustment.

By considering dual constraints jointly, DUET achieves a better balance between stability and plasticity. Extensive experiments on multiple continual learning benchmarks demonstrate that DUET consistently outperforms existing unidirectional projection methods and achieves state-of-the-art performance on most benchmarks. Our main contributions are summarized as follows:
\begin{itemize}
    \item We propose \textbf{DUET}, a bidirectional Jacobian null-space projection algorithm that jointly considers previous-task preservation and new-task adaptation through dual constraints, with theoretical guarantees on the existence of a feasible solution.
    
    \item We identify the limitation of existing cosine criteria for step-size selection under low effective-rank Jacobians and propose an effective-rank aware cosine criterion for more reliable adaptive projection.
    
    \item We conduct extensive experiments on multiple datasets, demonstrating that DUET consistently outperforms existing unidirectional projection methods and achieves state-of-the-art performance on most metrics.

\end{itemize}

\section{Related Works}
\label{sec:formulation_related_works}

A central challenge in continual learning is to acquire new knowledge without forgetting previously learned tasks. In PEFT paradigm, Low-Rank Adaptation (LoRA) is widely adopted~\cite{hu2021lora}. It freezes pretrained parameters $\theta_{t-1}$ from the previous task and parameterizes task-specific updates as low-rank matrices $\Delta\theta_t$, resulting in adapted parameters $\theta_{t-1} + \Delta\theta_t$, which can be naturally compatible with other anti-forgetting methods~\cite{wang2023orthogonal, chen2024coin}. 

\paragraph{Optimization Formulation.}
\label{sec:formulations}
Continual learning is often formulated as a constrained optimization problem~\cite{fang2025alphaedit, Paz2017GEM, Arslan2019AGEM, Tekmen2025IGEM}:
\begin{equation}
\begin{gathered}
\min_{\Delta\theta_t} \, L_{\mathcal{D}_t}(\theta_{t-1} + \Delta\theta_t) \\
\text{s.t.} \quad
L_{\mathcal{D}_{<t}}(\theta_{t-1} + \Delta\theta_t)
=
L_{\mathcal{D}_{<t}}(\theta_{t-1}).
\end{gathered}
\label{eq:constrained_opt}
\end{equation}
where $\mathcal{D}_t$ denotes the dataset of task $t$ and $\mathcal{D}_{<t} = \bigcup_{i=1}^{t-1}\mathcal{D}_i$ denote datasets of previous tasks. 

\paragraph{Surrogate Constraints.}
Since directly enforcing the loss preservation constraint is intractable and historical data cannot be fully retained, existing approaches adopt surrogate constraints, including subspace orthogonality and parameter-space orthogonality.

Subspace orthogonality enforce updates to be orthogonal to historical activation features $\Delta W_t X_{<t} = 0$, where $\Delta W_t$ denotes the layer-wise weight update matrix and $X_{<t}$ denotes representative features extracted from historical activations~\cite{saha2021gradient, liang2023adaptive, liang2024inflora}. It ensures unchanged layer outputs, which is overly restrictive and limits plasticity~\cite{wang2024milora, meng2024pissa, yang2024corda}.

Parameter-space orthogonality ensures first-order loss invariance. Specifically, performing a first-order Taylor expansion yields:
\begin{equation}
\label{eq:taylor_expansion}
\begin{aligned}
& L_{\mathcal{D}_{<t}}(\theta_{t-1} + \Delta\theta_t) = L_{\mathcal{D}_{<t}}(\theta_{t-1}) \\
& \quad + \langle \nabla_{\theta_{t-1}} L_{\mathcal{D}_{<t}}(\theta_{t-1}), \Delta\theta_t \rangle + \mathcal{O}(\|\Delta\theta_t\|^2),
\end{aligned}
\end{equation}
where $\nabla_{\theta_{t-1}} L_{\mathcal{D}_{<t}}(\theta_{t-1})$ denotes the gradient of the prior task loss, computed from a small batch of retained historical samples. Neglecting the higher-order residual $\mathcal{O}(\|\Delta\theta_t\|^2)$, preserving historical performance requires parameter-space gradient orthogonality~\cite{farajtabar2020orthogonal, liu2026octopus, zheng2026past}: $\langle \nabla_{\theta_{t-1}} L_{\mathcal{D}_{<t}}(\theta_{t-1}), \Delta\theta_t \rangle = 0$. Parameter-space orthogonality is far less restrictive and can be effectively enforced by projecting the post-hoc parameter update $\Delta\theta_t$ onto the Jacobian null space of previous tasks, detailed in Sec.~\ref{sec:Jacobian_null_space_proj}.

DUET adopts parameter-space orthogonality as its surrogate constraint to maximize optimization freedom for acquiring new tasks.

\paragraph{Solving the Constrained Optimization Problem}
\label{sec:solvers}
Existing methods either relax constraints into penalized losses via soft regularization~\cite{kirkpatrick2017overcoming, krukowski2025intact, liu2026octopus}, or strictly enforce them via projection across different training stages~\cite{hou2026continual}: pre-training orthogonal initialization~\cite{tang2026put, liang2024inflora, meng2024pissa, yang2024corda, wang2024milora}, during-training gradient projection~\cite{kirkpatrick2017overcoming, farajtabar2020orthogonal, krukowski2025intact, zhan2025bicriteria}, or post-hoc parameter rectification~\cite{ilharco2022editing, yadav2023ties, huang2025mitigating, vasani2025phlora, zheng2026past}.

Among these, post-hoc parameter projection achieves remarkable efficiency by enabling unconstrained adaptation on the new task before adjusting parameters, where Jacobian null-space projection method, epitomized by JANUS~\cite{zheng2026past}, achieves the most prominent performance. Following this promising paradigm, DUET advances unidirectional post-hoc projection into an alternating bidirectional projection scheme.
\section{Methods}
\label{sec:methods}
We propose DUET, an alternating bidirectional projection algorithm to solve the distance minimization problem, and further introduce an effective-rank aware cosine criterion to enable reliable adaptive step size selection.

\subsection{Distance Minimizing Formulation}
\label{sec:Distance_mini}
DUET introduces a novel formulation for continual learning via dual-point distance minimization:
\begin{equation}
\begin{gathered}
\min_{\theta_1, \theta_2} \|\theta_1 - \theta_2\|^2 \\
\text{s.t.} \quad L_{\mathcal{D}_{<t}}(\theta_1) = L_{\mathcal{D}_{<t}}(\theta_{t-1}), \quad L_{\mathcal{D}_t}(\theta_2) = L_{\mathcal{D}_t}(\hat{\theta}_t)
\end{gathered}
\label{eq:DUET}
\end{equation}
Unconstrained fine-tuning on task $t$ yields the candidate parameters $\hat \theta_t = \theta_{t-1} + \arg\min_{\Delta\theta_t} L_{\mathcal{D}_t}(\theta_{t-1} + \Delta\theta_t)$, which excel on the new task but suffer from catastrophic forgetting on prior tasks. While existing unidirectional projection methods only utilizes a mini-batch historical data $\mathcal{D}_{<t}$ for unidirectional projection, DUET maintains mini-batches from both $\mathcal{D}_{<t}$ and $\mathcal{D}_t$ to solve a constrained distance minimization problem.

Notably, existing unidirectional post-hoc projection methods can be viewed as a degenerate case of both Eqs.~\eqref{eq:constrained_opt} and~\eqref{eq:DUET}. Specifically, by fixing $\theta_2$ in Eq.~\eqref{eq:DUET} to the unconstrained optimum $\hat\theta_t $ acts as a parameter-distance proxy for $\min L_{\mathcal{D}_t}$ in Eq.~\eqref{eq:constrained_opt}

\subsection{Alternating Bidirectional Projection}
\label{sec:DUET_arch}
DUET solves the distance minimization in Eq.~\eqref{eq:DUET} via alternating Jacobian null-space projections.

\paragraph{Layer-wise Jacobian and Null-Space Projection.}
\label{sec:Jacobian_null_space_proj}
Following JANUS~\cite{zheng2026past}, parameter projection is executed layer-by-layer. For the $l$-th layer with weight matrix $W^{(l)} \in \mathbb{R}^{d_{\text{out}} \times d_{\text{in}}}$ and target displacement $\Delta W^{(l)}$, we denote their vectorized forms as $\theta^{(l)} = \operatorname{vec}(W^{(l)}) \in \mathbb{R}^{d_{\text{out}} d_{\text{in}}  \times 1}$ and $\Delta\theta^{(l)} = \operatorname{vec}(\Delta W^{(l)})\in \mathbb{R}^{d_{\text{out}} d_{\text{in}}  \times 1}$.

To instantiate the first-order loss invariance from Eq.~\eqref{eq:taylor_expansion} across each data sample, preserving performance on a dataset $\mathcal{D}$ with $m$ samples requires the projected parameter update $\Delta\theta^{(l)}_{\text{proj}}$ to be orthogonal to each sample's gradient vector, satisfying $\langle \operatorname{vec}(\nabla_{W^{(l)}} \ell_i), \Delta\theta^{(l)}_{\text{proj}} \rangle = 0$ for all $i \in \{1, \dots, m\}$, where $\ell_i$ denotes the loss on the $i$-th data sample and $\operatorname{vec}(\nabla_{W^{(l)}} \ell_i) \in \mathbb{R}^{d_{\text{out}} d_{\text{in}}  \times 1}$. Stacking these $m$ sample-wise gradient vectors row-wise compactly converts these individual conditions into the matrix orthogonality equation $J^{(l)} \Delta\theta^{(l)}_{\text{proj}} = \mathbf{0}$, where the layer Jacobian matrix $J^{(l)}$ is defined as:
\begin{equation}
J^{(l)} = \begin{bmatrix}
\operatorname{vec}\big(\nabla_{W^{(l)}} \ell_1 \big)^\top \\
\vdots \\
\operatorname{vec}\big(\nabla_{W^{(l)}} \ell_m \big)^\top
\end{bmatrix} \in \mathbb{R}^{m \times d_{\text{out}} d_{\text{in}}},
\label{eq:jacobian_def}
\end{equation}

To enforce this condition, any desired parameter displacement $\Delta\theta^{(l)}$ is projected onto the null space of $J^{(l)}$ via the orthogonal projection operator $\Pi^{(l)} = I - (J^{(l)})^\top (J^{(l)} (J^{(l)})^\top)^{-1} J^{(l)}$:
\begin{equation}
\Delta\theta^{(l)}_{\text{proj}} = \Pi^{(l)} \Delta\theta^{(l)}.
\label{eq:proj_formula}
\end{equation}
This projection directly ensures first-order loss preservation, as $ J^{(l)}\Delta\theta^{(l)}_{\text{proj}} = J^{(l)}\Pi^{(l)} \Delta\theta^{(l)} = \mathbf{0}.$

\begin{algorithm}[h]
\caption{DUET}
\label{alg:DUET}
\begin{algorithmic}[1]
\renewcommand{\algorithmicrequire}{\textbf{Input:}}
\renewcommand{\algorithmicensure}{\textbf{Output:}}
\Require Prior parameters $\boldsymbol{\theta}_{t-1}$, fine-tuned parameters $\hat{\boldsymbol{\theta}}_t$, exemplars $\mathcal{D}_{<t}, \mathcal{D}_t$, max rounds $K$.
\Ensure Final parameters $\boldsymbol{\theta}_t$.
\State Initialize dual parameters: $\boldsymbol{\theta}_1 \leftarrow \boldsymbol{\theta}_{t-1}, \quad \boldsymbol{\theta}_2 \leftarrow \hat{\boldsymbol{\theta}}_t$.
\For{$k = 1, \dots, K$}
  \For{each phase $(\boldsymbol{\theta}, \boldsymbol{\theta}_{\text{target}}, \tilde{\mathcal{D}}) \in$ \hfill\break
       \null\hspace{1.2em} $\{(\boldsymbol{\theta}_1, \boldsymbol{\theta}_2, \mathcal{D}_{<t}), \, (\boldsymbol{\theta}_2, \boldsymbol{\theta}_1, \mathcal{D}_t)\}$}
    \State Compute $J^{(l)}$ on $\tilde{\mathcal{D}}$ via Eq.~\eqref{eq:jacobian_def}
    \State $\Pi^{(l)} \leftarrow I - (J^{(l)})^\top \big(J^{(l)} (J^{(l)})^\top\big)^{-1} J^{(l)}$
    \State Compute $\Delta\boldsymbol{\theta}_{\text{proj}}^{(l)} \leftarrow \Pi^{(l)}(\boldsymbol{\theta}_{\text{target}}^{(l)} - \boldsymbol{\theta}^{(l)})$
    \State Initialize trial step size: $\alpha \leftarrow 1$
    \While{True}
      \State Let $\boldsymbol{\theta}_{\text{trial}} \leftarrow \boldsymbol{\theta} + \alpha \Delta\boldsymbol{\theta}_{\text{proj}}$
      \State Compute rank-aware cosine $\bar{c}_r$ via Eq.~\eqref{eq:DUET_cosine}
      \If{$\bar{c}_r \ge \tau$}
        \State \textbf{break}
      \EndIf
      \State $\alpha \leftarrow \beta \cdot \alpha$
    \EndWhile
    \If{$L_{\tilde{\mathcal{D}}}(\boldsymbol{\theta}_{\text{trial}}) \le L_{\tilde{\mathcal{D}}}(\boldsymbol{\theta})$}
      \State Update parameter: $\boldsymbol{\theta} \leftarrow \boldsymbol{\theta}_{\text{trial}}$
    \EndIf
  \EndFor
  \If{neither $\boldsymbol{\theta}_1$ nor $\boldsymbol{\theta}_2$ updated}
    \State \textbf{break}
  \EndIf
\EndFor
\State \Return $\boldsymbol{\theta}_t \leftarrow \arg\min_{\boldsymbol{\theta} \in \{\boldsymbol{\theta}_1, \boldsymbol{\theta}_2\}} \big( L_{\mathcal{D}_{<t}}(\boldsymbol{\theta}) + L_{\mathcal{D}_t}(\boldsymbol{\theta}) \big)$
\end{algorithmic}
\end{algorithm}

\paragraph{Alternating Bidirectional Projection Mechanism.}
As illustrated in Fig.~\ref{fig:DUET_mechanism}, DUET resolves the parameter distance minimization problem in Eq.~\eqref{eq:DUET} via alternating bidirectional projection between two parameter states: $\theta_1$, initialized at $\theta_{t-1}$, is constrained to preserve the historical task loss $L_{\mathcal{D}_{<t}}$, while $\theta_2$, initialized at the unconstrained optimum $\hat\theta_t$, is constrained to preserve the new-task loss $L_{\mathcal{D}_t}$.

Instead of dealing with a computationally prohibitive joint constrained problem directly, DUET solves the optimization problem by iteratively contracting the distance $\|\theta_1 - \theta_2\|$ along the loss-preserving null space of each task. 

Specifically, this alternating scheme operates by taking turns fixing one parameter as the target $\theta_{\text{target}}$ while updating the other moving parameter $\theta$ toward it within its feasible null space. During each round $r \in \{1, \dots, R\}$, the total displacement $\Delta\theta = \theta_{\text{target}} - \theta$ is orthogonally decomposed into the permissible null-space update $\Delta\theta_{\text{proj}} = \Pi \Delta\theta$ and its gradient complement $\Delta\theta_{\perp} = (I - \Pi)\Delta\theta$. Because the first-order approximation holds only locally, parameters are updated iteratively with an adaptive step size $\alpha \in (0, 1]$:
\begin{equation}
\theta^{(l)} \leftarrow \theta^{(l)} + \alpha \Delta\theta^{(l)}_{\text{proj}},
\label{eq:step_update}
\end{equation}
where the criterion for adaptively determining $\alpha$ will be elaborated in Sec.~\ref{sec:step_size}. 

This target-directed update rule is executed across two complementary phases in each round, as outlined in Algorithm~\ref{alg:DUET}. In the \textbf{Old-to-New Phase} ($\theta_1 \to \theta_2$), $\theta_2$ is fixed as the target $\theta_{\text{target}} = \theta_2$, and $\theta_1$ updates via $\Pi_{<t}$ constructed from prior data samples $\mathcal{D}_{<t}$ (Eqs.~\eqref{eq:jacobian_def}--\eqref{eq:proj_formula}); because $J^{(l)}_{<t} \Delta\theta^{(l)}_{\text{proj}} = 0$, it preserves prior-task performance. Symmetrically, in the \textbf{New-to-Old Phase} ($\theta_2 \to \theta_1$), $\theta_1$ is fixed as the target $\theta_{\text{target}} = \theta_1$, and $\theta_2$ updates via $\Pi_t$ constructed from current data samples $\mathcal{D}_t$, where $J^{(l)}_{t} \Delta\theta^{(l)}_{\text{proj}} = 0$ guarantees new-task adaptation not degraded. By alternating these two phases across rounds, both parameters continuously contract the gap toward the shared intersection of their respective preservation manifolds.

\paragraph{Termination Criteria.}
The termination of the alternating optimization is directly governed by the acceptance of candidate updates. While the step size $\alpha$ is determined independently via a loss-agnostic subspace criterion (Sec.~\ref{sec:step_size}), a candidate update is accepted to proceed if and only if its corresponding task loss after projection does not exceed the loss prior to this step. Specifically, an update to $\theta_1$ is accepted if $L_{\mathcal{D}_{<t}}(\theta_1 + \alpha\Delta\theta_{\text{proj}}) \le L_{\mathcal{D}_{<t}}(\theta_1)$, while an update to $\theta_2$ is accepted if $L_{\mathcal{D}_t}(\theta_2 + \alpha\Delta\theta_{\text{proj}}) \le L_{\mathcal{D}_t}(\theta_2)$.

The alternating optimization terminates when neither parameter can proceed or when a maximum round budget is reached. If an update on one side violates this loss threshold and is rejected (e.g., updating $\theta_1$ toward $\theta_2$), DUET does not stop immediately; instead, subsequent iterations are dedicated solely to updating the opposite parameter ($\theta_2$ toward $\theta_1$) until updates on both sides are rejected. In practice, the process is mostly terminated by reaching a preset round cap, at which point the parameter distance $\|\theta_1 - \theta_2\|$ typically contracts to $10^{-2}$--$10^{-3}$ of its initial value (see Fig.~\ref{fig:param_dist}).

Upon termination, parameter with lower joint loss is output as the deployed model:
\begin{equation}
\theta_t = \arg\min_{\theta \in \{\theta_1, \theta_2\}} \left[ L_{\mathcal{D}_{<t}}(\theta) + L_{\mathcal{D}_t}(\theta) \right].
\label{eq:model_selection}
\end{equation}

\subsection{Adaptive Step Size via Cosine Similarity}
\label{sec:step_size}

To determine an admissible step size $\alpha$ in Eq.~\eqref{eq:step_update}, we assess whether local linearity holds by measuring the geometric distortion of the gradient subspace before and after a candidate update. Starting from $\alpha = 1.0$, the step is accepted if the subspace alignment satisfies the cosine similarity criterion; otherwise, $\alpha$ is decayed ($\alpha \leftarrow \beta\alpha$) and re-evaluated.

Through singular value decomposition (SVD), let $J = U\Sigma V^\top$ and $J' = U'\Sigma'{V'}^\top$ with $J, J' \in \mathbb{R}^{m \times d_{\text{out}} d_{\text{in}}}$. Because parameter updates reside in the parameter space $\mathbb{R}^{d_{\text{out}} d_{\text{in}}}$ rather than the output space $\mathbb{R}^m$, we consider the right-singular vectors $V, V' \in \mathbb{R}^{d_{\text{out}} d_{\text{in}} \times m}$ (instead of the output-space bases $U, U'$), whose columns form orthonormal bases spanning the respective gradient subspaces in the parameter space. A straightforward measure of subspace alignment averages the cosines of all $m$ principal angles~\cite{zheng2026past}:
\begin{equation}
\bar{c} = \frac{1}{m} \sum_{j=1}^m \sigma_j\big(V^\top V'\big).
\label{eq:raw_cosine}
\end{equation}
However, this measure can be misleading when the Jacobian $J$ is effective-rank deficient ($r < m$), where the remaining $m - r$ noise-dominated directions are near-orthogonal, causing trailing singular values to collapse toward zero. In such cases, the cosine similarity is systematically underestimated—remaining low even at a zero step size ($\alpha = 0$) where $J$ is identical before and after projection—which falsely triggers step rejection, over-shrinks $\alpha$, and stalls optimization (see Fig.~\ref{fig:cosine_criterion}). To resolve this limitation, we determine the effective rank $r = \sum_{i=1}^m \mathbb{I}(\lambda_i > \epsilon \lambda_1)$ from the Gram matrix $G = JJ^\top \in \mathbb{R}^{m \times m}$ (with cutoff $\epsilon = 10^{-3}$), where $\lambda_1$ is the largest eigenvalue of $G$. We then evaluate alignment solely across these $r$ informative directions:
\begin{equation}
\bar{c}_r = \frac{1}{r} \sum_{j=1}^r \sigma_j\big(V_r^\top V_r'\big).
\label{eq:DUET_cosine}
\end{equation}
where $V_r, V'_r \in \mathbb{R}^{d_{\text{out}} d_{\text{in}} \times r}$ denote the submatrices formed by the leading $r$ columns of $V$ and $V'$, respectively. A candidate step is accepted once $\bar{c}_r \ge \tau$ (e.g., $\tau = 0.95$).

\paragraph{Efficient Implementation.}
To mitigate the space complexity of the algorithm, we adopt ghost projection and ghost orientation comparison techniques as JANUS~\cite{zheng2026past}.

\subsection{Convergence Analysis of Alternating Projection}
\label{sec:proof}

We establish that the alternating dual-point projection strictly contracts and converges to zero distance.

Let $\theta$ be the moving point and $\theta_{\text{target}}$ be the fixed target, with displacement $\Delta\theta = \theta_{\text{target}} - \theta$. The null-space projection matrix $\Pi = I - J^\top (J J^\top)^{+} J$ is symmetric ($\Pi^\top = \Pi$) and idempotent ($\Pi^2 = \Pi$), where layer superscript $(l)$ is omitted for brevity. Decomposing $\Delta\theta$ into the projected component $\Delta\theta_{\text{proj}} = \Pi \Delta\theta$ and the orthogonal complement $\Delta\theta_{\perp} = (I - \Pi)\Delta\theta$, their inner product satisfies:
\begin{equation}
\Delta\theta_{\text{proj}}^\top \Delta\theta_{\perp} = \Delta\theta^\top \Pi^\top (I - \Pi) \Delta\theta = \Delta\theta^\top (\Pi - \Pi^2) \Delta\theta = 0.
\end{equation}
By the Pythagorean theorem, the initial squared distance expands as:
\begin{equation}
\|\Delta\theta\|^2 = \|\Delta\theta_{\text{proj}}\|^2 + \|\Delta\theta_{\perp}\|^2.
\label{eq:pythagorean_proof}
\end{equation}

When updating with step size $\alpha \in (0, 1]$ along the projected direction, the parameter becomes $\theta' = \theta + \alpha \Delta\theta_{\text{proj}}$, resulting in the updated residual $\Delta\theta' = \theta_{\text{target}} - \theta' = (1 - \alpha)\Delta\theta_{\text{proj}} + \Delta\theta_{\perp}$. Applying the orthogonality from Eq.~\eqref{eq:pythagorean_proof}, the norm of the updated error vector satisfies:
\begin{align}
\|\Delta\theta'\|^2 &= (1 - \alpha)^2 \|\Delta\theta_{\text{proj}}\|^2 + \|\Delta\theta_{\perp}\|^2 \notag \\
&= \|\Delta\theta\|^2 - \alpha(2 - \alpha) \|\Delta\theta_{\text{proj}}\|^2 < \|\Delta\theta\|^2.
\label{eq:step_strictly_smaller}
\end{align}
Because $\alpha(2 - \alpha) > 0$ for all $\alpha \in (0, 1]$, Eq.~\eqref{eq:step_strictly_smaller} strictly guarantees that $\|\Delta\theta'\| < \|\Delta\theta\|$ whenever $\Delta\theta_{\text{proj}} \neq 0$,  with the contraction constant $\gamma$ bounded by:
\begin{equation}
\gamma = \sup \frac{\|\Delta\theta'\|}{\|\Delta\theta\|} = \sqrt{1 - \alpha(2 - \alpha) \min \frac{\|\Delta\theta_{\text{proj}}\|^2}{\|\Delta\theta\|^2}}.
\end{equation}

\begin{figure}[h]
    \centering
    \includegraphics[width=1\linewidth]{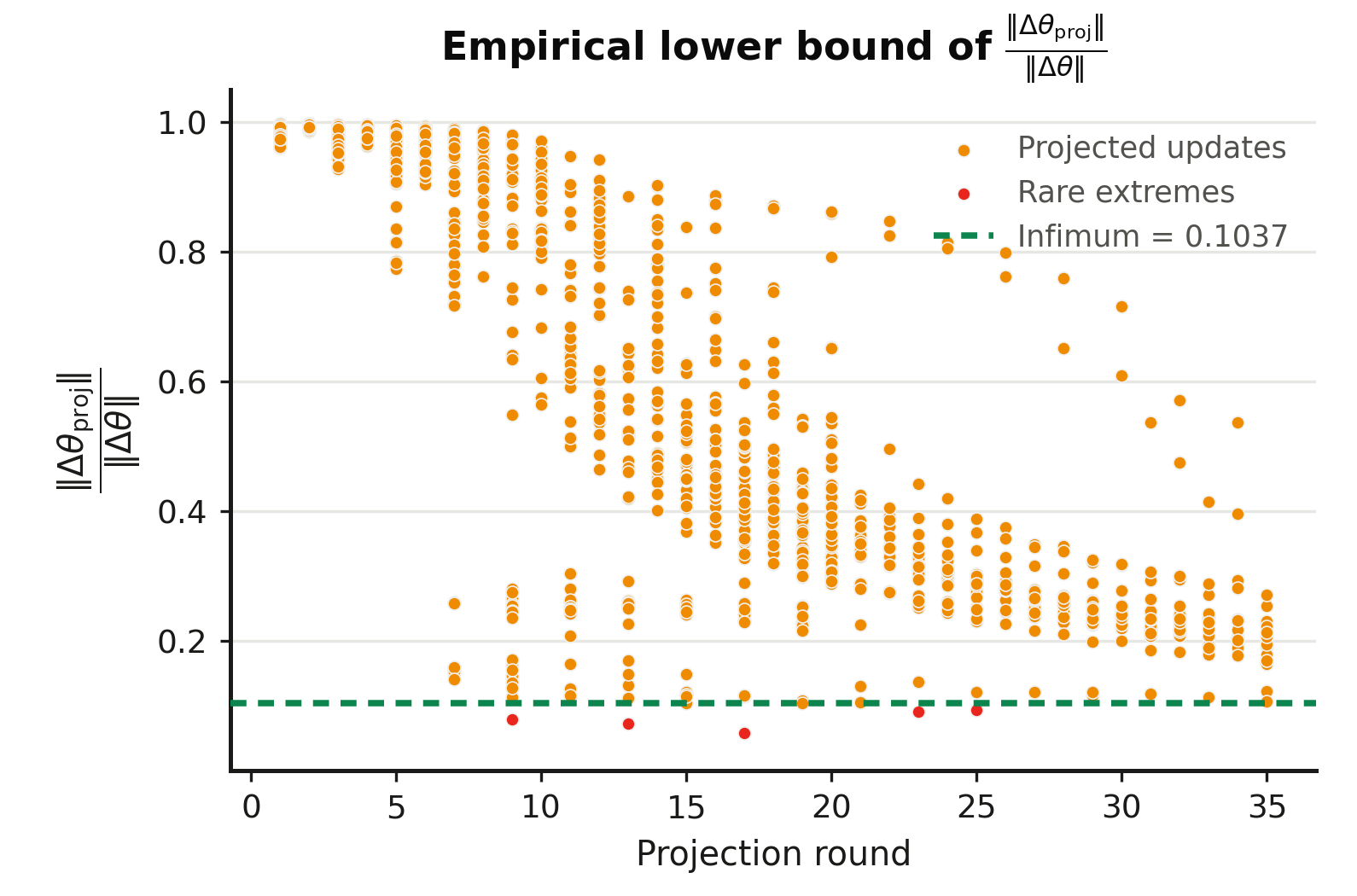}
    \caption{Null-space projection ratio $\frac{\|\Delta\theta_{\text{proj}}\|}{\|\Delta\theta\|}$ across rounds. The ratio remains strictly positive ($\ge 0.1$), empirically validating the contraction property ($\gamma < 1$).}
    \label{fig:contraction}
\end{figure}

\begin{table*}[t]
\centering
\small
\caption{\textbf{Continual learning performance on ImageNet-R and DomainNet.} The best results are highlighted in \textbf{bold}.}
\label{tab:main_results1}
\resizebox{\textwidth}{!}{
\setlength{\tabcolsep}{4.5pt}
\begin{tabular}{lcccccc}
\toprule
{\textbf{Method}} & \multicolumn{4}{c}{\textbf{ImageNet-R} (10 Tasks)} & \multicolumn{2}{c}{\textbf{DomainNet} (5 Tasks)} \\
\cmidrule(lr){2-5} \cmidrule(lr){6-7}
& $ACC_5$ ($\uparrow$) & $\overline{ACC}_5$ ($\uparrow$) & $ACC_{10}$ ($\uparrow$) & $\overline{ACC}_{10}$ ($\uparrow$) & $ACC_5$ ($\uparrow$) & $\overline{ACC}_5$ ($\uparrow$) \\
\midrule
Joint & -- & -- & $75.22 \pm 0.44$ & -- & $74.69 \pm 0.11$ & -- \\
Seqential~\cite{hu2021lora} & $74.43 \pm 1.06$ & $82.19 \pm 0.49$ & $58.82 \pm 1.22$ & $72.05 \pm 0.13$ & $70.76 \pm 0.24$ & $78.30 \pm 0.05$ \\
\midrule
CODA-P~\cite{smith2023coda} & $77.11 \pm 0.72$ & $80.71 \pm 0.40$ & $72.62 \pm 0.34$ & $77.28 \pm 0.46$ & $73.07 \pm 0.06$ & $78.50 \pm 0.04$ \\
LoRA-Null v1~\cite{tang2026put} & $74.33 \pm 0.96$ & $82.31 \pm 0.66$ & $54.83 \pm 0.60$ & $71.41 \pm 0.66$ & $70.57 \pm 0.07$ & $78.17 \pm 0.08$ \\
LoRA-Null v2~\cite{tang2026put} & $70.38 \pm 0.35$ & $76.55 \pm 0.41$ & $66.29 \pm 0.53$ & $72.04 \pm 0.45$ & $71.59 \pm 0.10$ & $77.32 \pm 0.06$ \\
PEPG~\cite{qiao2024gradient} & $77.46 \pm 0.50$ & $83.76 \pm 0.36$ & $71.05 \pm 0.44$ & $77.70 \pm 0.37$ & $72.68 \pm 0.11$ & $79.00 \pm 0.05$ \\
InfLoRA~\cite{liang2024inflora} & $80.45 \pm 0.43$ & $85.23 \pm 0.28$ & $73.96 \pm 0.47$ & $80.43 \pm 0.21$ & $72.97 \pm 0.10$ & $79.04 \pm 0.07$ \\
JANUS~\cite{zheng2026past} & $79.06 \pm 0.51$ & $84.76 \pm 0.44$ & $72.83 \pm 0.36$ & $79.75 \pm 0.36$ & $72.12 \pm 0.11$ & $78.93 \pm 0.04$ \\
\midrule
\textbf{DUET (Ours)} & $\mathbf{82.25 \pm 0.49}$ & $\mathbf{86.74 \pm 0.28}$ & $\mathbf{75.48 \pm 0.39}$ & $\mathbf{82.12 \pm 0.36}$ & $\mathbf{73.32 \pm 0.12}$ & $\mathbf{79.53 \pm 0.06}$ \\
\bottomrule
\end{tabular}
}
\end{table*}

\begin{figure*}
    \centering
    \includegraphics[width=\textwidth]{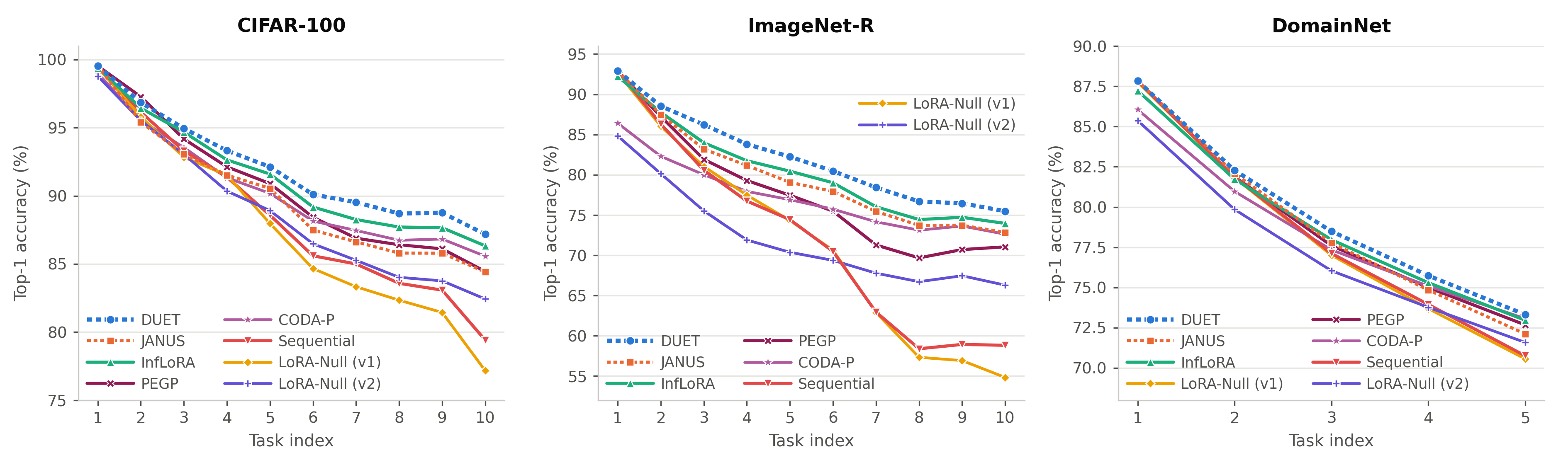}
    \caption{Top-1 accuracy after each incremental task on CIFAR-100, DomainNet, and ImageNet-R.}
    \label{fig:acc_all_datasets}
\end{figure*}

As shown in Fig.~\ref{fig:contraction}, the empirical minimum of the null-space projection ratio is approximately $0.1$:
\begin{equation}
\min \frac{\|\Delta\theta_{\text{proj}}\|}{\|\Delta\theta\|} \approx 0.1
\end{equation}
indicating that the projection retains a non-negligible component of the original displacement throughout iterations.

By the \textbf{Banach Fixed-Point Theorem (Contraction mapping Theorem)}, iteratively applying this contractive mapping guarantees that the sequence of parameter differences contracts geometrically:
\begin{equation}
\|\theta_1^{(k)} - \theta_2^{(k)}\| \le \gamma^k \|\theta_{t-1} - \hat \theta_t\|.
\end{equation}
As $k \to \infty$, the distance converges uniquely to zero:
\begin{equation}
\label{eq:convergence_limit}
\lim_{k \to \infty} \|\theta_1^{(k)} - \theta_2^{(k)}\| = 0.
\end{equation}

\section{Experiments}
\label{sec:experiments}

\subsection{Benchmarks and Baselines}
\label{sec:exp_setup}

\noindent\textbf{Datasets.} We evaluate DUET on three standard continual learning benchmarks~\cite{gao2023unified, liang2024inflora}:

\begin{itemize}[leftmargin=*]
    \item \textbf{CIFAR-100}~\cite{krizhevsky2009learning}: 50{,}000 training and 10{,}000 test images over 100 classes, split into 10 class-incremental tasks (10 classes per task). This is a class-incremental benchmark with relatively small image resolution (32$\times$32).
    \item \textbf{ImageNet-R}~\cite{hendrycks2021many}: 23{,}767 training and 5{,}769 test images from 200 ImageNet-derived classes, divided into 10 tasks (20 classes per task). This is a class-incremental benchmark with high intra-class variation.
    \item \textbf{DomainNet}~\cite{peng2019moment}: 338{,}804 training and 69{,}089 test images over 345 classes, organized into 5 domain-incremental tasks (one domain per task, 69 shared classes). This is a domain-incremental benchmark covering 5 distinct graphic domains.
\end{itemize}

\noindent\textbf{Baselines.} We compare DUET against multiple continual learning methods:
For \textit{Joint} method, a single LoRA module is optimized over all tasks simultaneously. For other baselines, a new LoRA module is added for each new task. Among them, \textit{Sequential}~\cite{hu2021lora} serves as the naive baseline trained without any anti-forgetting constraints, while the others introduce different previous-tasks protection mechanisms: two null-space initialization variants \textit{LoRA-Null v1/v2}~\cite{tang2026put} (v1 merely initializes LoRA parameters in the null space of previous tasks; v2 initializes and freezes the matrix $B$ to be orthogonal to previous tasks while only updating the expansion matrix $A$ where $\Delta W=AB$), the interference-free projection method \textit{InfLoRA}~\cite{liang2024inflora}, \textit{PEGP}~\cite{qiao2024gradient} which projects gradients onto the orthogonal subspace of past feature representations, the post-hoc projection approach \textit{JANUS}~\cite{zheng2026past}, and \textit{CODA-P}~\cite{smith2023coda}, as a representative rehearsal-free prompt-tuning approach.

Notably, JANUS baseline is equipped with our proposed effective-rank aware cosine criterion; otherwise, the original criterion perpetually rejects all step sizes and fails to execute even a single projection from the first round (see Sec.~\ref{sec:exp_cosine}).

\subsection{Main Results and Quantitative Comparison}
\label{sec:main_results}

Following standard CL evaluation protocols~\cite{gao2023unified, liang2024inflora}, we evaluate all methods with top-1 accuracy after learning task $k$ ($ACC_k$) and the average incremental accuracy up to task $k$ ($\overline{ACC}_k = \frac{1}{k}\sum_{i=1}^k ACC_i$). Specifically, $ACC_k$ denotes the accuracy evaluated across all seen tasks up to task $k$, and top-1 accuracy requires the ground-truth label to be the single highest-ranked prediction. For multi-run experiments, we report the mean and sample standard deviation across 5 independent random seeds.

Table~\ref{tab:main_results1},~\ref{tab:main_results2} presents the quantitative comparison across three benchmarks. Fig.~\ref{fig:acc_all_datasets} illustrates the $ACC_k$ top-1 accuracy after each incremental task on the benchmarks. As shown, DUET consistently outperforms JANUS~\cite{zheng2026past} and other baselines throughout the entire learning process, rather than only at the final stage, demonstrating consistent superiority throughout the entire learning process. It even matches the joint training performance on ImageNet-R.

\begin{table}[h]
\centering
\small
\caption{\textbf{Continual learning performance on CIFAR-100.}}
\label{tab:main_results2}
\resizebox{\columnwidth}{!}{
\setlength{\tabcolsep}{3pt}
\begin{tabular}{lcccc}
\toprule
{\textbf{Method}} & \multicolumn{4}{c}{\textbf{CIFAR-100} (10 Tasks)} \\
\cmidrule(lr){2-5}
& $ACC_5$ ($\uparrow$) & $\overline{ACC}_5$ ($\uparrow$) & $ACC_{10}$ ($\uparrow$) & $\overline{ACC}_{10}$ ($\uparrow$) \\
\midrule
Joint & -- & -- & $88.04 \pm 0.52$ & -- \\
Sequential & $88.58 \pm 1.08$ & $93.79 \pm 0.36$ & $79.41 \pm 1.36$ & $88.56 \pm 0.55$ \\
\midrule
CODA-P & $90.18 \pm 0.32$ & $93.92 \pm 0.24$ & $85.57 \pm 0.55$ & $90.43 \pm 0.22$ \\
LoRA-Null v1 & $87.97 \pm 0.59$ & $93.51 \pm 0.35$ & $77.17 \pm 2.42$ & $87.65 \pm 0.82$ \\
LoRA-Null v2 & $88.92 \pm 0.22$ & $93.32 \pm 0.06$ & $82.43 \pm 0.50$ & $88.86 \pm 0.26$ \\
PEPG & $90.90 \pm 0.73$ & $94.80 \pm 0.30$ & $84.45 \pm 0.84$ & $90.63 \pm 0.40$ \\
InfLoRA & $91.60 \pm 0.39$ & $94.95 \pm 0.22$ & $86.33 \pm 0.21$ & $91.39 \pm 0.15$ \\
JANUS & $90.55 \pm 0.26$ & $94.01 \pm 0.39$ & $84.41 \pm 0.35$ & $90.02 \pm 0.20$ \\
\midrule
\textbf{DUET}& $\mathbf{92.13 \pm 0.28}$ & $\mathbf{95.36 \pm 0.09}$ & $\mathbf{87.18 \pm 0.36}$ & $\mathbf{92.11 \pm 0.17}$ \\
\bottomrule
\end{tabular}
}
\end{table}

\paragraph{Varying Pre-trained Model}
In addition to the ImageNet experiments using ViT-B/16~\cite{dosovitskiy2021image} weights supervisedly pretrained on ImageNet-21K (Sup-21K) in Table~\ref{tab:main_results1}, we further evaluate algorithms with ViT-B/16 models pretrained in different ways, including DINO~\cite{caron2021emerging} and iBOT~\cite{zhou2022ibot} as reported in Table~\ref{tab:main_results_ibot_dino}.

DUET consistently outperforms JANUS on both backbones. On DINO, InfLoRA~\cite{liang2024inflora} achieves higher final accuracy, while DUET attains comparable average accuracy (75.63\% vs. 75.98\%). A potential reason is that DINO relies solely on the global classification (\text{CLS}) token during pre-training, unlike iBOT and Sup-21K. Since both approaches attempt to preserve all tokens at both the \text{CLS} and patch levels, this distinction in pre-training objectives may account for the slight performance gap.

\begin{table}[h]
\centering
\small
\caption{\textbf{Continual learning performance on iBOT and DINO.}}
\label{tab:main_results_ibot_dino}

\resizebox{\columnwidth}{!}{
\setlength{\tabcolsep}{3pt}
\begin{tabular}{lcccc}
\toprule
{\textbf{Method}} & \multicolumn{4}{c}{\textbf{iBOT}} \\
\cmidrule(lr){2-5}
& $ACC_5$ ($\uparrow$) & $\overline{ACC}_5$ ($\uparrow$) & $ACC_{10}$ ($\uparrow$) & $\overline{ACC}_{10}$ ($\uparrow$) \\
\midrule
Joint & -- & -- & $72.56 \pm 0.33$ & -- \\
Sequential & $73.28 \pm 0.71$ & $79.87 \pm 0.47$ & $55.06 \pm 1.83$ & $69.77 \pm 0.62$ \\
\midrule
CODA-P & $77.07 \pm 0.51$ & $81.03 \pm 0.19$ & $68.74 \pm 0.71$ & $76.35 \pm 0.31$ \\
LoRA-Null v1 & $72.30 \pm 0.59$ & $79.70 \pm 0.43$ & $48.21 \pm 2.64$ & $67.44 \pm 0.27$ \\
LoRA-Null v2 & $63.80 \pm 0.37$ & $70.07 \pm 0.14$ & $56.79 \pm 0.44$ & $64.78 \pm 0.20$ \\
PEGP & $76.93 \pm 0.78$ & $81.86 \pm 0.19$ & $67.24 \pm 0.43$ & $75.72 \pm 0.17$ \\
InfLoRA & $79.19 \pm 0.32$ & $82.89 \pm 0.26$ & $69.92 \pm 0.54$ & $77.76 \pm 0.19$ \\
JANUS & $77.39 \pm 0.36$ & $82.54 \pm 0.39$ & $66.62 \pm 0.37$ & $76.63 \pm 0.33$ \\
\midrule
\textbf{DUET}& $\mathbf{80.08 \pm 0.20}$ & $\mathbf{84.08 \pm 0.42}$ & $\mathbf{70.67 \pm 0.55}$ & $\mathbf{79.05 \pm 0.37}$ \\
\bottomrule
\end{tabular}
}

\vspace{2mm}

\resizebox{\columnwidth}{!}{
\setlength{\tabcolsep}{3pt}
\begin{tabular}{lcccc}
\toprule
{\textbf{Method}} & \multicolumn{4}{c}{\textbf{DINO}} \\
\cmidrule(lr){2-5}
& $ACC_5$ ($\uparrow$) & $\overline{ACC}_5$ ($\uparrow$) & $ACC_{10}$ ($\uparrow$) & $\overline{ACC}_{10}$ ($\uparrow$) \\
\midrule
Joint & -- & -- & $69.09 \pm 0.34$ & -- \\
Sequential & $68.12 \pm 0.96$ & $76.23 \pm 0.51$ & $50.61 \pm 1.02$ & $65.52 \pm 0.40$ \\
\midrule
CODA-P & $72.54 \pm 0.43$ & $77.61 \pm 0.43$ & $64.97 \pm 0.35$ & $72.76 \pm 0.36$ \\
LoRA-Null v1 & $67.21 \pm 1.20$ & $76.15 \pm 0.54$ & $44.67 \pm 2.49$ & $63.89 \pm 0.54$ \\
LoRA-Null v2 & $62.24 \pm 0.81$ & $68.13 \pm 0.53$ & $54.67 \pm 0.14$ & $62.84 \pm 0.42$ \\
PEPG & $72.31 \pm 0.49$ & $78.33 \pm 0.29$ & $61.75 \pm 0.48$ & $71.43 \pm 0.38$ \\
InfLoRA & $\mathbf{77.12 \pm 0.39}$ & $81.18 \pm 0.25$ & $\mathbf{67.06 \pm 0.38}$ & $\mathbf{75.98 \pm 0.15}$ \\
JANUS & $73.63 \pm 0.67$ & $79.47 \pm 0.44$ & $61.72 \pm 0.70$ & $72.70 \pm 0.43$ \\
\midrule
\textbf{DUET}& $76.53 \pm 0.55$ & $\mathbf{81.52 \pm 0.36}$ & $65.90 \pm 0.48$ & $75.63 \pm 0.33$ \\
\bottomrule
\end{tabular}
}
\end{table}

\subsection{Validation of Effective-Rank Cosine Criterion}
\label{sec:exp_cosine}

\begin{figure}[h]
    \centering
    \includegraphics[width=0.95\linewidth]{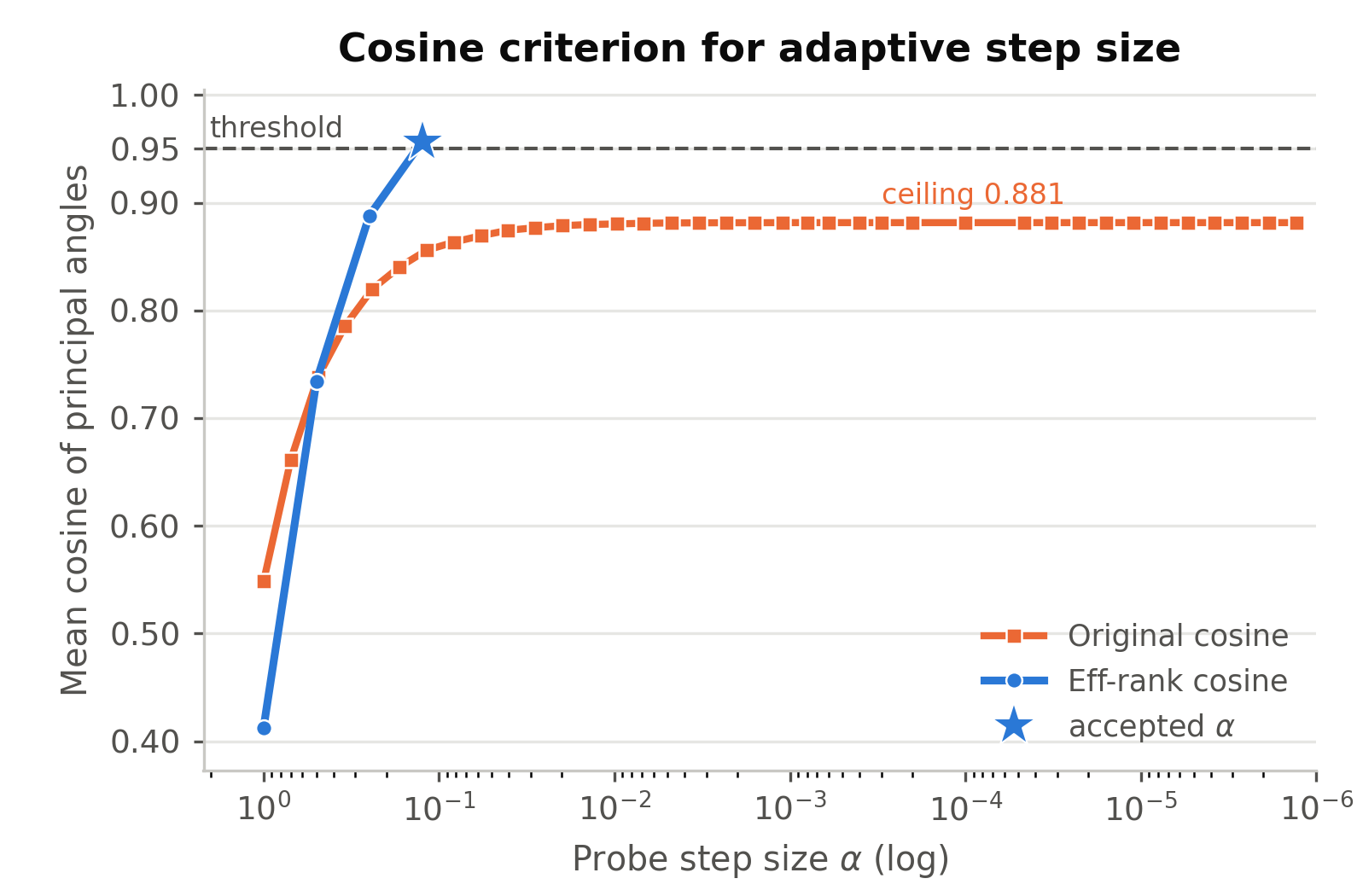}
    \caption{Step-size adaptation under different cosine criteria.}
    \label{fig:cosine_criterion}
\end{figure}

Fig.~\ref{fig:cosine_criterion} verifies the failure mode of the standard cosine criterion during step-size probing. Due to rank-deficient Jacobians, the original cosine in JANUS~\cite{zheng2026past} exhibits an artificial ceiling at $0.881$, perpetually falling short of the threshold $\tau = 0.95$ and rejecting updates even as $\alpha \to 10^{-6}$. In contrast, by filtering out uninformative noise directions, our effective-rank cosine successfully crosses the threshold and accepts an admissible step size, effectively preventing premature optimization stagnation.

\subsection{Parameter Convergence}
\label{sec:param_convergence}

At projection round $k$, the relative distance between $\theta_1$ and $\theta_2$ across all layers is defined as:
\begin{equation}
\label{eq:param_dist}
\Delta{D_{12}}^{(k)} = \frac{\|\theta_1^{(k)} - \theta_2^{(k)}\|}{\|\theta_{t-1} - \hat{\theta}_t\|} = \frac{\left( \sum_l \|\theta_1^{(k, l)} - \theta_2^{(k, l)}\|_F^2 \right)^{1/2}}{\left( \sum_l \|\theta_{t-1}^{(l)} - \hat{\theta}_t^{(l)}\|_F^2 \right)^{1/2}}.
\end{equation}
While Eq.~\eqref{eq:convergence_limit} establishes asymptotic convergence to zero, in practice, reducing the distance to $\|\theta_1^{(k)} - \theta_2^{(k)}\| < 10^{-2} \|\theta_{t-1}-\hat{\theta}_t\|$ suffices for effective convergence, given the finite floating-point precision of neural networks.

\begin{figure}[h]
    \centering
    \includegraphics[width=0.98\linewidth]{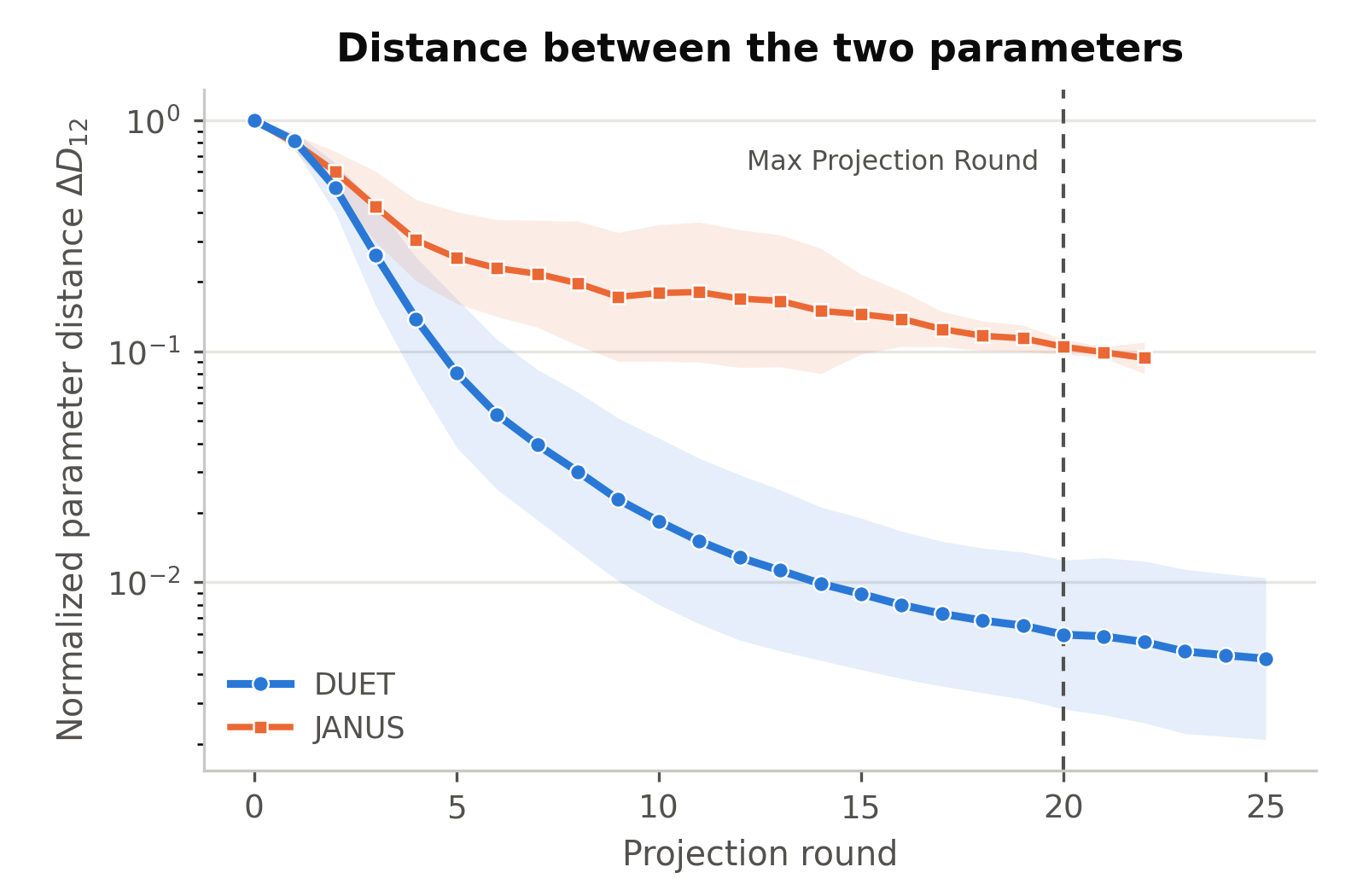}
    \caption{Normalized parameter distance across projection rounds.}
    \label{fig:param_dist}
\end{figure}

We evaluate parameter convergence by tracking this normalized distance across projection rounds (Fig.~\ref{fig:param_dist}), where shaded areas denote $\pm 1$ standard deviation across seeds and tasks computed in log space. While JANUS plateaus around $10^{-1}$, DUET demonstrates a consistently faster decline, reducing the distance below $10^{-2}$. 

\subsection{Implementation Details}
\label{sec:implementation_details}

Across all datasets, we use a ViT-B/16 backbone~\cite{dosovitskiy2021image} pre-trained on ImageNet-21K and optimize with Adam. LoRA adapters are injected into the query, key, and value ($Q, K, V$) projection matrices of all multi-head self-attention layers, with other training configurations and the classification head architecture strictly following InfLoRA~\cite{liang2024inflora}. For memory efficiency, we follow JANUS~\cite{zheng2026past} and conduct both ghost projection and ghost orientation comparison techniques. To construct the historical Jacobian null-space constraints, DUET maintains an exemplar buffer of $128$ samples per task on CIFAR-100, $256$ on ImageNet, and $512$ on DomainNet. To control the memory footprint in long task sequences, the global historical memory budget is capped at a maximum of $2{,}048$ samples across all datasets, which is dynamically partitioned evenly among all prior tasks once exceeded. For a fair comparison, our implemented JANUS~\cite{zheng2026past} and LoRA-Null~\cite{tang2026put}, strictly adopt the exact same training hyperparameters and historical exemplar allocation budgets. InfLoRA~\cite{liang2024inflora} computes a task's activation matrix via a single forward pass over its full training set after training on each task, following original setting.

For parameter projection, we set the initial step size $\alpha = 1.0$, decay factor $\beta = 0.5$, alignment threshold $\tau = 0.95$, and spectral truncation cutoff $\epsilon = 10^{-3}$. All experiments across datasets and seeds are executed on a single NVIDIA A100 GPU.
\section{Conclusions}
\label{sec:conclusions}

In this paper, we present DUET, a bidirectional Jacobian null-space projection framework for continual learning that jointly addresses previous-task preservation and new-task adaptation. DUET formulates post-hoc parameter adjustment as a constrained optimization problem that minimizes the distance between the parameters before and after new-task training, subject to Jacobian null-space constraints from both previous and new tasks. We solve this problem through alternating projections and theoretically show that the parameter distance contracts toward zero. Furthermore, we identify the limitations of existing cosine-based step-size criteria under rank-deficient Jacobians and introduce an effective-rank-aware criterion for more reliable adaptive projection. Extensive experiments on CIFAR-100, ImageNet-R, and DomainNet demonstrate that DUET outperforms existing unidirectional projection methods and achieves state-of-the-art performance on most evaluation metrics. These results highlight the benefits of jointly considering previous-task preservation and new-task adaptation in post-hoc parameter projection for continual learning.

\paragraph{Limitations.}
DUET relies on first-order Taylor approximations to enforce loss preservation. Due to their local validity, multi-step projections and cosine-based step adaptations inevitably accumulate approximation errors, indicating that loss invariance cannot be perfectly guaranteed in practice.
\section*{LLM Usage Statement}

GPT-6 Instant and Gemini-3.8 Flash are used to assist with grammar correction, paper revising and improving sentence flow across the manuscript. Claude Opus-5.5 is used to assist in code writing and debugging. Generative AI tools have negligible involvement in research ideation, experimental design, or figure generation. All generated code and text were reviewed and verified by the authors, who take full responsibility for the contents of this paper.

{
    \small
    \bibliographystyle{ieeenat_fullname}
    \bibliography{main}

@String(PAMI  = {IEEE TPAMI})

@String(CVPR  = {CVPR})

@String(ICCV  = {ICCV})

@String(NIPS  = {NeurIPS})

@String(ICLR  = {ICLR})

@String(AAAI = {AAAI})

@String(TNNLS = {IEEE TNNLS})

@String(AISTATS={AISTATS})

@String(NAACL={NAACL})

@inproceedings{zheng2026past,
title     = {Past, Future, All at Once: Breaking Stability-Plasticity Dilemma via Post-hoc {JANUS} Rectification},
author    = {Zheng, Zhilong and Tao, Letian and Guan, Yang and Yang, Yujie and Xiong, Wei and Sheng, Kehua and Zhang, Bo and Duan, Jingliang and Li, Keqiang and Li, Shengbo Eben},
booktitle = NIPS,
year={2026}
}

@inproceedings{liang2024inflora,
  title={{InfLoRA}: Interference-free Low-rank Adaptation for Continual Learning},
  author={Liang, Yan-Shuo and Li, Wu-Jun},
  booktitle=CVPR,
  pages={23638--23647},
  year={2024}
}

@inproceedings{tang2026put,
  title={Put the Space of {LoRA} Initialization to the Extreme to Preserve Pre-trained Knowledge},
  author={Tang, Pengwei and Hu, Xiaolin and Liu, Yong and Ding, Lizhong and Zhang, Dongjie and Wu, Xing and Zhang, Debing},
  booktitle=AAAI,
  pages={33232--33240},
  year={2026}
}

@inproceedings{ilharco2022editing,
  title={Editing Models with Task Arithmetic},
  author={Ilharco, Gabriel and Ribeiro, Marco Tulio and Wortsman, Mitchell and Schmidt, Ludwig and Hajishirzi, Hannaneh and Farhadi, Ali},
  booktitle=ICLR,
  year={2023}
}

@inproceedings{yadav2023ties,
  title={{TIES}-merging: Resolving Interference When Merging Models},
  author={Yadav, Prateek and Tam, Derek and Choshen, Leshem and Raffel, Colin A and Bansal, Mohit},
  booktitle=NIPS,
  pages={7093--7115},
  year={2023}
}

@article{zhan2025bicriteria,
  title={Bicriteria Policy Optimization for High-Accuracy Reinforcement Learning},
  author={Zhan, Guojian and Zhang, Xiangteng and Zhang, Feihong and Tao, Letian and Li, Shengbo Eben},
  journal=TNNLS, 
  volume  = {37},
  number  = {1},
  pages   = {312--326},
  year    = {2026}
}

@inproceedings{liang2023adaptive,
  title={Adaptive Plasticity Improvement for Continual Learning},
  author={Liang, Yan-Shuo and Li, Wu-Jun},
  booktitle=CVPR,
  pages={7816--7825},
  year={2023}
}

@inproceedings{saha2021gradient,
  title={Gradient Projection Memory for Continual Learning},
  author={Saha, Gobinda and Garg, Isha and Roy, Kaushik},
  booktitle=ICLR,
  year={2021}
}

@inproceedings{hu2021lora,
  title={{LoRA}: Low-Rank Adaptation of Large Language Models},
  author={Hu, Edward J. and Shen, Yelong and Wallis, Phillip and
          Allen-Zhu, Zeyuan and Li, Yuanzhi and Wang, Shean and
          Wang, Lu and Chen, Weizhu},
  booktitle=ICLR,
  year={2022}
}

@inproceedings{vasani2025phlora,
  title={{PHLoRA}: Data-free Post-hoc Low-Rank Adapter Extraction from Full-rank Checkpoint},
  author={Vasani, Bhoomit and FitzGerald, Jack and Fang, Anjie and Vaish, Sushmit},
  booktitle={Proceedings of IJCNLP-AACL},
  pages={1992--1999},
  year={2025}
}

@article{hou2026continual,
  title={Continual Learning in Transition},
  author={Hou, Zhiyan and Zhang, Dan and Feng, Tao and Wang, Liyuan and Li, Wei and Hao, Xiangzhao and An, Hongyan and Fang, Junfeng and Ma, Haokai and Xu, Zhaohui and others},
  journal={arXiv preprint arXiv:2608.06216},
  year={2026}
}

@article{krukowski2025intact,
  title={{InTAct}: Interval-based Task Activation Consolidation for Continual Learning},
  author={Krukowski, Patryk and Miksa, Jan and Helm, Piotr and Tabor, Jacek and Spurek, Przemys{\l}aw and others},
  journal={arXiv preprint arXiv:2511.17439},
  year={2025}
}

@inproceedings{liu2026octopus,
  title={Octopus: History-Free Gradient Orthogonalization for Continual Learning in Multimodal Large Language Models},
  author={Liu, Yuehao and Guan, Shanyan and Zhang, Weijia and Shang, Xuanming and Ge, Yanhao and Li, Wei and Ma, Chao},
  booktitle=CVPR,
  year={2026}
}

@inproceedings{wang2023orthogonal,
  title={Orthogonal Subspace Learning for Language Model Continual Learning},
  author={Wang, Xiao and Chen, Tianze and Ge, Qiming and Xia, Han and Bao, Rong and Zheng, Rui and Zhang, Qi and Gui, Tao and Huang, Xuan-Jing},
  booktitle={Findings of EMNLP},
  pages={10658--10671},
  year={2023}
}

@inproceedings{farajtabar2020orthogonal,
  title={Orthogonal Gradient Descent for Continual Learning},
  author={Farajtabar, Mehrdad and Azizan, Navid and Mott, Alex and Li, Ang},
  booktitle=AISTATS,
  pages={3762--3773},
  year={2020}
}

@inproceedings{chen2024coin,
  title={{Coin}: A benchmark of Continual Instruction Tuning for Multimodel Large Language Models},
  author={Chen, Cheng and Zhu, Junchen and Luo, Xu and Shen, Heng T and Song, Jingkuan and Gao, Lianli},
  booktitle=NIPS,
  pages={57817--57840},
  year={2024}
}

@inproceedings{wang2024milora,
    title = "{M}i{L}o{RA}: Harnessing Minor Singular Components for Parameter-Efficient {LLM} Finetuning",
    author = "Wang, Hanqing and Li, Yixia and Wang, Shuo and Chen, Guanhua and Chen, Yun",
    booktitle = NAACL,
    year={2025},
    pages={4823--4836},
}

@inproceedings{meng2024pissa,
  title={Pissa: Principal singular values and singular vectors adaptation of large language models},
  author={Meng, Fanxu and Wang, Zhaohui and Zhang, Muhan},
  booktitle=NIPS,
  pages={121038--121072},
  year={2024}
}

@inproceedings{yang2024corda,
  title={{CorDA}: Context-oriented Decomposition Adaptation of Large Language Models for Task-aware Parameter-efficient Fine-tuning},
  author={Yang, Yibo and Li, Xiaojie and Zhou, Zhongzhu and Song, Shuaiwen L and Wu, Jianlong and Nie, Liqiang and Ghanem, Bernard},
  booktitle=NIPS,
  pages={71768--71791},
  year={2024}
}

@inproceedings{huang2025mitigating,
  title={Mitigating Catastrophic Forgetting in Large Language Models with Forgetting-aware Pruning},
  author={Huang, Wei and Cheng, Anda and Wang, Yinggui},
  booktitle=EMNLP,
  pages={21853--21867},
  year={2025}
}

@article{kirkpatrick2017overcoming,
  title   = {Overcoming Catastrophic Forgetting in Neural Networks},
  author  = {Kirkpatrick, James and Pascanu, Razvan and Rabinowitz, Neil and Veness, Joel and Desjardins, Guillaume and Rusu, Andrei A and Milan, Kieran and Quan, John and Ramalho, Tiago and Grabska-Barwinska, Agnieszka and others},
  journal = {Proceedings of the National Academy of Sciences},
  volume  = {114},
  number  = {13},
  pages   = {3521--3526},
  year    = {2017}
}

@inproceedings{gao2023unified,
  title     = {A Unified Continual Learning Framework with General Parameter-Efficient Tuning},
  author    = {Gao, Qiankun and Zhao, Chen and Sun, Yifan and Xi, Teng and Zhang, Gang and Ghanem, Bernard and Zhang, Jian},
  booktitle = ICCV,
  pages     = {11449--11459},
  year      = {2023}
}

@inproceedings{hendrycks2021many,
  title     = {The Many Faces of Robustness: A Critical Analysis of Out-of-Distribution Generalization},
  author    = {Hendrycks, Dan and Basart, Steven and Mu, Norman and Kadavath, Saurav and Wang, Frank and Dorundo, Evan and Desai, Rahul and Zhu, Tyler and Parajuli, Samyak and Guo, Mike and Song, Dawn and Steinhardt, Jacob and Gilmer, Justin},
  booktitle = ICCV,
  pages     = {8340--8349},
  year      = {2021}
}

@techreport{krizhevsky2009learning,
  title       = {Learning Multiple Layers of Features from Tiny Images},
  author      = {Krizhevsky, Alex and Hinton, Geoffrey},
  institution = {University of Toronto},
  year        = {2009}
}

@inproceedings{peng2019moment,
  title     = {Moment Matching for Multi-Source Domain Adaptation},
  author    = {Peng, Xingchao and Bai, Qinxun and Xia, Xide and Huang, Zijun and Saenko, Kate and Wang, Bo},
  booktitle = ICCV,
  pages     = {1406--1415},
  year      = {2019}
}

@inproceedings{smith2023coda,
    title     = {{CODA-Prompt}: {COntinual} Decomposed Attention-Based Prompting for Rehearsal-Free Continual Learning},
    author    = {Smith, James Seale and Karlinsky, Leonid and Gutta, Vyshnavi and Cascante-Bonilla, Paola and Kim, Donghyun and Arbelle, Assaf and Panda, Rameswar and Feris, Rogerio and Kira, Zsolt},
    booktitle = CVPR,
    year      = {2023},
    pages     = {11909--11919}
}

@article{qiao2024gradient,
  title   = {Gradient Projection for Continual Parameter-Efficient Tuning},
  author  = {Qiao, Jingyang and Zhang, Zhizhong and Tan, Xin and Qu, Yanyun and Zhang, Wensheng and Han, Zhi and Xie, Yuan},
  journal = PAMI,
  volume  = {47},
  number  = {10},
  pages   = {9316--9329},
  year    = {2025}
}

@inproceedings{caron2021emerging,
  title     = {Emerging Properties in Self-Supervised Vision Transformers},
  author    = {Caron, Mathilde and Touvron, Hugo and Misra, Ishan and Jegou, Herv{\'e} and Mairal, Julien and Bojanowski, Piotr and Joulin, Armand},
  booktitle = ICCV,
  pages     = {9630--9640},
  year      = {2021}
}

@inproceedings{zhou2022ibot,
  title     = {{iBOT}: Image {BERT} Pre-Training with Online Tokenizer},
  author    = {Zhou, Jinghao and Wei, Chen and Wang, Huiyu and Shen, Wei and Xie, Cihang and Yuille, Alan and Kong, Tao},
  booktitle = ICLR,
  year      = {2022}
}

@inproceedings{dosovitskiy2021image,
  title     = {An Image is Worth 16x16 Words: Transformers for Image Recognition at Scale},
  author    = {Dosovitskiy, Alexey and Beyer, Lucas and Kolesnikov, Alexander and Weissenborn, Dirk and Zhai, Xiaohua and Unterthiner, Thomas and Dehghani, Mostafa and Minderer, Matthias and Heigold, Georg and Gelly, Sylvain and Uszkoreit, Jakob and Houlsby, Neil},
  booktitle = ICLR,
  year      = {2021}
}

@article{PARISI201954,
title = {Continual lifelong learning with neural networks: A review},
journal = {Neural Networks},
volume = {113},
pages = {54--71},
year = {2019},
author = {German I. Parisi and Ronald Kemker and Jose L. Part and Christopher Kanan and Stefan Wermter}
}

@inproceedings{kemker2018measuring,
  title={Measuring catastrophic forgetting in neural networks},
  author={Kemker, Ronald and McClure, Marc and Abitino, Angelina and Hayes, Tyler and Kanan, Christopher},
  booktitle=AAAI,
  volume={32},
  number={1},
  year={2018}
}

@book{doi:10.1137/1.9781421407944,
author = {Golub, Gene H. and Van Loan, Charles F.},
title = {Matrix Computations},
edition={4th},
year = {2013},
publisher={SIAM},
doi = {10.1137/1.9781421407944}
}

@book{doi:10.1137/1.9780898719697,
author = {Hansen, Per Christian},
title = {Rank-Deficient and Discrete Ill-Posed Problems},
year = {1998},
publisher={SIAM},
doi = {10.1137/1.9780898719697}
}

@inproceedings{luo2026keeplora,
  title={Keeplora: Continual Learning with Residual Gradient Adaptation},
  author={Luo, Mao-Lin and Zhou, Zi-Hao and Zhang, Yi-Lin and Wan, Yuanyu and Zhang, Min-Ling and Wei, Tong},
  booktitle=ICLR,
  pages={136017--136040},
  year={2026}
}

@inproceedings{hsu2024safelora,
 author = {Hsu, Chia-Yi and Tsai, Yu-Lin and Lin, Chih-Hsun and Chen, Pin-Yu and Yu, Chia-Mu and Huang, Chun-Ying},
 booktitle = NIPS,
 title = {Safe LoRA: The Silver Lining of Reducing Safety Risks when Finetuning Large Language Models},
 pages = {65072--65094},
 year = {2024}
}

@inproceedings{fang2025alphaedit,
 author = {Fang, Junfeng and Jiang, Houcheng and Wang, Kun and Ma, Yunshan and Shi, Jie and Wang, Xiang and He, Xiangnan and Chua, Tat-Seng},
 booktitle = ICLR,
 title = {AlphaEdit: Null-Space Constrained Knowledge Editing for Language Models},
 year = {2025}
}

@inproceedings{Paz2017GEM,
 author = {Lopez-Paz, David and Ranzato, Marc\textquotesingle Aurelio},
 booktitle = NIPS,
 title = {Gradient Episodic Memory for Continual Learning},
 year = {2017}
}

@inproceedings{Arslan2019AGEM,
  author       = {Arslan Chaudhry and Marc'Aurelio Ranzato and Marcus Rohrbach and Mohamed Elhoseiny},
  title        = {Efficient Lifelong Learning with {A-GEM}},
  booktitle    = ICLR,
  year         = {2019}
}

@inproceedings{Tekmen2025IGEM,
  title     = {{{GEM}-Style} Constraints for {PEFT} with Dual Gradient Projection in {LoRA}},
  author    = {Tekmen, Brian and Yin, Jason and Tong, Qianqian},
  booktitle = {IEEE International Conference on Data Mining Workshops (ICDMW)},
  pages     = {2736--2743},
  year      = {2025}
}

@inproceedings{SVDLLM,
 author = {Wang, Xin and Zheng, Yu and Wan, Zhongwei and Zhang, Mi},
 booktitle = ICLR,
 title = {SVD-LLM: Truncation-aware Singular Value Decomposition for Large Language Model Compression},
 pages = {19299--19319},
 year = {2025}
}

@inproceedings{SVDQuant,
 author = {Li, Muyang and Lin, Yujun and Zhang, Zhekai and Cai, Tianle and Li, Xiuyu and Guo, Junxian and Xie, Enze and Meng, Chenlin and Zhu, Jun-Yan and Han, Song},
 booktitle = ICLR,
 title = {SVDQuant: Absorbing Outliers by Low-Rank Component for 4-Bit Diffusion Models},
 pages = {97831--97858},
 year = {2025}
}

@inproceedings{DobiSVD,
 author = {Wang, Qinsi and Ke, Jinghan and Tomizuka, Masayoshi and Keutzer, Kurt and Xu, Chenfeng},
 booktitle = ICLR,
 title = {Dobi-SVD: Differentiable SVD for LLM Compression and Some New Perspectives},
 pages = {12561--12590},
 year = {2025}
}

@article{guan2026enhanced,
  title={Enhanced Integrated Decision and Control for High-Level Automated Vehicles and Its Experiment Verification},
  author={Guan, Yang and Tang, Liye and Lyu, Yao and Li, Shengbo Eben and Yang, Yujie and Sheng, Kehua and Zhang, Bo and Li, Keqiang},
  journal={IEEE Transactions on Automation Science and Engineering},
  year={2026}
}

@article{guan2027beyond,
  title={Beyond Rule-based Systems: Generalized Nudge Decision-making for Autonomous Robo-delivery via Multi-task Transformer Learning},
  author={Guan, Yang and Xiao, Liming and Huang, Bai and Yu, Guangyuan and Li, Xiao and Wang, Naizheng and Duan, Jingliang and Li, Shengbo Eben},
  journal={Transportation Research Part C: Emerging Technologies},
  volume={194},
  pages={105998},
  year={2027}
}

@article{guan2021direct,
  title={Direct and Indirect Reinforcement Learning},
  author={Guan, Yang and Li, Shengbo Eben and Duan, Jingliang and Li, Jie and Ren, Yangang and Sun, Qi and Cheng, Bo},
  journal={International Journal of Intelligent Systems},
  volume={36},
  number={8},
  pages={4439--4467},
  year={2021}
}

@article{CVCL,
  title   = {Class-Incremental Learning: Survey and Performance Evaluation on Image Classification},
  author  = {Masana, Marc and Liu, Xialei and Twardowski, Bart{\l}omiej and Menta, Mikel and Bagdanov, Andrew D. and van de Weijer, Joost},
  journal = PAMI,
  volume  = {45},
  number  = {5},
  pages   = {5513--5533},
  year    = {2023}
}

@article{LLMCL,
  title   = {Continual Learning of Large Language Models: A Comprehensive Survey},
  author  = {Shi, Haizhou and Xu, Zihao and Wang, Hengyi and Qin, Weiyi and Wang, Wenyuan and Wang, Yibin and Wang, Zifeng and Ebrahimi, Sayna and Wang, Hao},
  journal = {ACM Computing Surveys},
  volume  = {58},
  number  = {5},
  year    = {2025}
}
}


\end{document}